\documentclass{article} 
\usepackage{iclr2015,times}
\usepackage{hyperref}
\usepackage{url}
\usepackage[margin=2.75cm]{geometry}
\usepackage{amsmath}
\usepackage{amssymb}
\usepackage{caption}
\usepackage{subcaption}
\usepackage{algorithm}
\usepackage{amsthm}

\usepackage[noend]{algpseudocode}
\usepackage[pdftex]{graphicx}

\newtheorem{innercustomthm}{Theorem}
\newenvironment{customthm}[1]
  {\renewcommand\theinnercustomthm{#1}\innercustomthm}
  {\endinnercustomthm}

\graphicspath{ {./images/} }

\title{Understanding Minimum Probability Flow for RBMs Under Various
  Kinds of Dynamics}

\author{
    Daniel Jiwoong Im \&  Ethan Buchman \& Graham W. Taylor\\
School of Engineering\\
University of Guelph\\
Guelph, On, Canada\\
\texttt{\{imj,ebuchman,gwtaylor\}@uoguelph.ca} \\
}

\iclrfinalcopy 

\begin{document}

\maketitle

\begin{abstract}
Energy-based models are popular in machine learning due to the elegance of their 
formulation and their relationship to statistical physics.  Among these, the 
Restricted Boltzmann Machine (RBM), and its staple training algorithm contrastive divergence (CD),
have been the prototype for some recent advancements in the unsupervised training of deep neural networks.  
However, CD has limited theoretical motivation, and can in some cases produce undesirable behavior.
Here, we investigate the performance of Minimum Probability Flow (MPF) learning for training RBMs. 
Unlike CD, with its focus on approximating an intractable partition function via Gibbs sampling, 
MPF proposes a tractable, consistent, objective function defined in terms of a Taylor expansion 
of the KL divergence with respect to sampling dynamics. 
Here we propose a more general form for the sampling dynamics in MPF, 
and explore the consequences of different choices for these dynamics for training RBMs. 
Experimental results show MPF outperforming CD for various RBM configurations.

\end{abstract}

\section{Introduction}
A common problem in machine learning is to estimate the parameters of a high-dimensional probabilistic model using gradient descent on the model's negative log likelihood.
For exponential models where $p(x)$ is proportional to the exponential
of a negative potential function $F(x)$, the gradient of the data
negative log-likelihood takes the form
\begin{equation}
    \nabla_{\theta} =
                \frac{1}{|\mathcal{D}|} \left( \sum_{x \in \mathcal{D}} \frac{\partial{F(x)}}{\partial{\theta}} -
                    \sum_{x} p(x) \frac{\partial{F(x)}}{\partial{\theta}} \right)
\end{equation}

where the sum in the first term is over the dataset, $\mathcal{D}$, and the sum in the second term is over the entire domain of $x$.
The first term has the effect of pushing the parameters in a direction that decreases the energy surface of the model at the training data points,
while the second term increases the energy of all possible states.
Since the second term is intractable for all
but trivial models, we cannot, in practice, accommodate for every state of $x$, but rather resort to sampling.
We call states in the sum in the first term {\em positive particles} and those in the second term {\em negative particles}, in accordance
with their effect on the likelihood (opposite their effect on the energy). 
Thus, the intractability of the second term becomes a problem of {\em negative particle selection} (NPS).

The most famous approach to NPS is Contrastive Divergence (CD) \citep{Hinton2002}, which is the centre-piece of unsupervised
neural network learning in energy-based models. ``CD-k'' proposes to sample the negative particles by applying a Markov chain Monte Carlo (MCMC)
transition operator $k$ times to each data state. This is in contrast to taking an unbiased sample from the distribution
by applying the MCMC operator a large number of times until the distribution reaches equilibrium, which is often prohibitive for practical applications.
Much research has attempted to better understand this approach and the reasoning behind its success or failure \citep{Sutskever2010, MacKay2001}, leading to many variations being proposed
from the perspective of improving the MCMC chain. Here, we take a more general approach to the problem of NPS, in particular, through the lens of the Minimum
Probability Flow (MPF) algorithm \citep{Sohl-Dickstein2011a}.

MPF works by introducing a continuous dynamical system over the model's distribution, 
such that the equilibrium state of the dynamical system is the distribution used to model the data.
The objective of learning is to minimize the flow of probability from data states to non-data states after infinitesimal evolution under the model's dynamics.
Inuitively, the less a data vector evolves under the dynamics, the closer it is to an equilibrium point; 
or from our perspective, the closer the equilibrium distribution is to the data.
In MPF, NPS is replaced by a more explicit notion of {\em connectivity} between states. Connected states are ones between which probability can flow under the dynamical system. Thus, rather than attempting to approximate an intractable function (as in CD-k), we run a simple optimization over an explicit, continuous dynamics,
and actually never have to run the dynamics themselves.

Interestingly, MPF and CD-k have gradients with remarkably similar form. 
In fact, the CD-k gradients can be seen as a special case of the MPF gradients - 
that is, MPF provides a generalized form which reduces to CD-k under a special dynamics.
Moreover, MPF provides a consistent estimator for the model parameters, while CD-k as typically formalized is an update heuristic,
that can sometimes do bizarre things like go in circles in parameter space \citep{MacKay2001}.
Thus, in one aspect, MPF {\em solves} the problem of contrastive divergence by reconceptualizing it as probability flow under an explicit
dynamics, rather than the convenient but biased sampling of an intractable function.
The challenge thus becomes one of how to design the dynamical system.

This paper makes the following contributions. First, we provide an explanation of MPF that begins from the familiar territory of CD-k, 
rather than the less familiar grounds of the master equation. While familiar to physicists, the master equation is an apparent obscurity in machine learning, 
due most likely to its general intractability. Part of the attractiveness of MPF is the way it circumvents that intractability. Second, we derive a generalized form 
for the MPF transition matrix, which defines the dynamical system. Third, we provide a Theano \citep{Theano} based implementation of MPF and a number
of variants of MPF that run efficiently on GPUs\footnote{https://github.com/jiwoongim/minimum\_probability\_flow\_learning}. Finally, we compare and contrast variants of MPF with those of CD-k, and experimentally demonstrate
that variants of MPF outperform CD-k for Restricted Boltzmann Machines
trained on MNIST and on Caltech-101.

\section{Restricted Boltzmann Machines}

While the learning methods we discuss apply to undirected probabilistic graphical models in general,
we will use the Restricted Boltzmann Machine (RBM) as a canonical example. An RBM is an undirected bipartite graph with visible (observed) variables
$\mathbf{v} \in \lbrace 0, 1\rbrace^D$ and
hidden (latent) variables $\mathbf{h} \in \lbrace 0,1\rbrace^H$ \citep{smolensky1986}. 
The RBM is an energy-based model where the energy of state $\mathbf{v ,h}$ is given by
\begin{equation}
    E(\mathbf{v},\mathbf{h};\theta) = - \sum_i\sum_j W_{ij}v_ih_j - \sum_i b_i v_i - \sum_j c_j h_j
\label{rbm_energy}
\end{equation}
where $\theta = \lbrace W, \mathbf{b, c} \rbrace$ are the parameters of the model.
The marginalized probability over visible variables is formulated from the Boltzmann distribution, 
\begin{align}
    \label{rbm_pv}
    p(\mathbf{v};\theta) &= \frac{p^*(\mathbf{v};\theta)}{Z(\theta)} = \frac{1}{Z(\theta)} \sum_\mathbf{h} \exp \Bigg(\frac{-1}{\tau}E(\mathbf{v},\mathbf{h};\theta) \Bigg)
\end{align}
such that $Z(\theta)= \sum_{\mathbf{v},\mathbf{h}}  \exp \big(\frac{-1}{\tau}E(\mathbf{v},\mathbf{h};\theta)\big)$ 
is a normalizing constant and $\tau$ is the thermodynamic temperature.
We can marginalize over the binary hidden states in Equation \ref{rbm_energy}
and re-express in terms of a new energy $F(\mathbf{v})$,
\begin{align}
    \label{rbm_pv_f}
    F(\mathbf{v};\theta) = -\log \sum_{\mathbf{h}} \exp \bigg(\frac{-1}{\tau}E(\mathbf{v},\mathbf{h})\bigg) &= \frac{1}{\tau}\sum_i^{D}v_i b_i - \frac{1}{\tau}\sum^{H}_{j=1} \log \Bigg(1+\exp \bigg(c_j + \sum_i^Dv_iW_{i,j} \bigg) \Bigg) \\
    p(\mathbf{v};\theta) &=\frac{ \exp \big(-F(\mathbf{v};\theta) \big)}{Z(\theta)} 
\end{align}
Following physics, this form of the energy is better known as a free energy, as it expresses the difference between the average energy and the entropy of a distribution, in this case, that of $p( \mathbf{h} | \mathbf{v} )$. Defining the distribution in terms of free energy as $p(\mathbf{v};\theta)$
is convenient since it naturally copes with the presence of latent variables.


The key characteristic of an RBM is the simplicity of inference due to conditional independence between visible and hidden states:
\begin{align*}
    \label{rbm_cond_prob}
    p(\mathbf{h}|\mathbf{v}) = \prod_j p(h_j|\mathbf{v})\text{, \indent } & \text{ }p(h_j=1|\mathbf{v}) = \sigma(\sum_i W_{ij}v_i + c_j)\\
    p(\mathbf{v}|\mathbf{h}) = \prod_i p(v_i|\mathbf{h})\text{, \indent } & \text{ }p(v_i=1|\mathbf{h}) = \sigma(\sum_j W_{ij}h_j + b_i)
\end{align*}
where $\sigma(z) = 1/(1+\exp{(-z}))$.

This leads naturally to a block Gibbs sampling dynamics, used universally for sampling from RBMs. Hence, in an RBM trained by CD-k, the connectivity (NPS) is determined with probability given by $k$ sequential block Gibbs sampling transitions.

We can formalize this by writing the learning updates for CD-k as follows
\begin{equation} 
    \label{eqn:cd_update}
    \Delta \theta_{CD-k} \propto - \sum_{j \in \mathcal{D}} \sum_{i \not\in \mathcal{D}}\Big( \frac{\partial F_j(\theta)}{\partial \theta}- \frac{\partial F_i(\theta)}{\partial \theta} \Big) T_{ij}
\end{equation}

where $T_{ij}$ is the probability of transitioning from state $j$ to state $i$ in $k$ steps of block Gibbs sampling. 
We can in principle replace $T_{ij}$ by any other transition operator, so long as it preserves the equilibrium distribution.
Indeed, this is what alternative methods, like Persistent CD \citep{Tieleman2009}, achieve.

\section{Minimum Probability Flow}
The key intuition behind MPF is that NPS can be reformulated in a firm theoretical context by treating the model distribution as the end point of some explicit continuous dynamics, and seeking to minimize the flow of probability away from the data under those dynamics. In this context then, NPS is no longer a sampling procedure employed to approximate an intractable function, but arises naturally out of the probability flow from data states to non-data states. 
That is, MPF provides a theoretical environment for the formal treatment of $T_{ij}$ that offers a much more general perspective of that operator than CD-k can. In the same vein, it better formalizes the notion of minimizing divergence between positive and negative particles.

\subsection{Dynamics of the Model}

The primary mathematical apparatus for MPF is a continuous time Markov chain known as the {\em master equation},
\begin{equation}
    \dot p_i = \sum_{j\neq i} [ \Gamma_{ij}p_{j}^{(t)} - \Gamma_{ji}p_{i}^{(t)} ]
\end{equation}

where $j$ are the data states and $i$ are the non-data states and $\Gamma_{ij}$ is the probability flow rate from state $j$ to state $i$.
Note that each state is a full vector of variables, and we are theoretically enumerating all states. $\dot p_i$ is the rate of change of the probability of state $i$, that is, the difference between the probability 
flowing out of any state $j$ into state $i$ and the probability 
flowing out of state $i$ to any other state $j$ at time $t$.
We can re-express $\dot p_i$ in a simple matrix form as
\begin{equation}
    \mathbf{\dot p} = \mathbf{\Gamma p}
\end{equation}
by setting $\Gamma_{ii}=-\sum_{i\neq j}\Gamma_{ji}p_{i}^{(t)}$. We note that if the transition matrix $\Gamma$ is ergodic, then the model has a unique stationary distribution. 

This is a common model for exploring statistical mechanical systems, but it is unwieldly in practice for two reasons, namely, the continuous time dynamics, and exponential size of the state space. For our purposes, we will actually find the former an advantage, and the latter irrelevant.

The objective of MPF is to minimize the KL divergence between the data distribution and the distribution after evolving an infinitesimal amount of time under the dynamics:
\begin{displaymath}
    \theta_{MPF} = \text{argmin}_\theta J(\theta), \text{  }
    J(\theta) = D_{KL} (p^{(0)}||p^{(\epsilon)}(\theta))
\end{displaymath}

Approximating $J(\theta)$ up to a first order Taylor expansion with respect to time $t$, our objective function reduces to 
\begin{equation}
    J(\theta) = \frac{\epsilon}{|\mathcal{D}|} \sum_{j\in \mathcal{D}}\sum_{i\not\in \mathcal{D}}\Gamma_{ij}
    \label{objective}
\end{equation}

and $\theta$ can be optimized by gradient descent on $J(\theta)$. Since $\Gamma_{ij}$ captures probability flow from state $j$ to state $i$, this objective function has the quite elegant interpretation of minimizing the probability flow from data states to non-data states \citep{Sohl-Dickstein2011a}.

\subsection{Form of the Transition Matrix}\label{sec:formTM}
MPF does not propose to {\em actually} simulate these dynamics. There is, in fact, no need to, as the problem formulation reduces to a rather simple optimization problem with no intractable component. However, we must provide a means for computing the matrix coefficients $\Gamma_{ij}$. Since our target distribution is the distribution defined by the RBM, we require $\Gamma$ to be a function of the energy, or more particularly, the parameters of the energy function.

A sufficient (but not necessary) means to guarantee that the distribution $\mathbf{p}^{\infty}\left(\theta\right)$ is a fixed point of the dynamics is to 
choose $\Gamma$ to satisfy detailed balance, that is
\begin{equation}
    \Gamma_{ji}p_{i}^{(\infty)} (\theta) = \Gamma_{ij}p_{j}^{(\infty)} (\theta).
\label{detail_balance}
\end{equation}

The following theorem provides a general form for the transition matrix such that the equilibrium distribution is that of the RBM:
\begin{customthm}{1} \label{theorem1}
    Suppose $p_j^{(\infty)}$ is the probability of state $j$ and $p_i^{(\infty)}$ is the probability of
    state $i$. Let the transition matrix be 
    \begin{equation}
        \Gamma_{ij} = g_{ij}\exp \left( \frac{o(F_i-F_j) +1}{2} (F_j-F_i) \right)
        \label{eqn:new_gamma}
    \end{equation}
    such that $o(\cdot)$ is any odd function, where $g_{ij}$ is the symmetric connectivity between the states $i$ and $j$.
    Then this transition matrix satisfies detailed balance in Equation \ref{detail_balance}.
\end{customthm}
The proof is provided in Appendix~\ref{app:appendexA}. The transition matrix proposed by \citep{Sohl-Dickstein2011a}
is thus the simplest case of Theorem 1, found by setting $o(\cdot) = 0$ and $g_{ij} = g_{ji}$:
\begin{equation}
    \Gamma_{ij} = g_{ij}\exp{\Big(\frac{1}{2}(F_j(\theta)-F_i(\theta)\Big)} .\
\end{equation}

Given a form for the transition matrix, we can now evaluate the gradient of $J(\theta)$
\begin{align*}
    \frac{\partial J(\theta)}{\partial \theta} & = 
        \frac{\epsilon}{|\mathcal{D}|} 
        \sum_{j\in \mathcal{D}}\sum_{i\not\in \mathcal{D}}
            \Big( \frac{\partial F_j(\theta)}{\partial \theta} 
                - \frac{\partial F_i(\theta)}{\partial \theta} \Big) 
                    T_{ij}  \\
    T_{ij} & = g_{ij}\exp\Big(\frac{1}{2}\big(F_j(\theta) - F_i(\theta)\big)\Big)
\end{align*}

\noindent and observe the similarity to the formulation given for the RBM trained by CD-k (Equation~\ref{eqn:cd_update}). 
Unlike with CD-k, however, this expression was derived through an explicit dynamics and well-formalized minimization objective.

\section{Probability Flow Rates $\Gamma$} \label{transition_rates}
At first glance, MPF might appear doomed, due to the size of $\Gamma$,
namely $2^D \times 2^D$, and the problem of enumerating all of the states.
However, the objective function in Equation \ref{objective} summing over the $\Gamma_{ij}$'s
only considers transitions between data states $j$ (limited in size by our data set)
and non-data states $i$ (limited by the sparseness of our design). By specifying  $\Gamma$
to be sparse, the intractability disappears, and complexity is dominated by the size of the
dataset.

Using traditional methods, an RBM can be trained in two ways, either with sampled negative particles, like
in CD-k or PCD (also known as stochastic maximum likelihood) \citep{Hinton2002,Tieleman2009}, or via an inductive
principle, with fixed sets of ``fantasy cases'', like in
general score matching, ratio matching, or pseudolikelihood \citep{Aapo2005, Marlin2011, Besag1975}.
In a similar manner, we can define $\Gamma$ by specifying the connectivity function $g_{ij}$ either as
a distribution from which to sample or as fixed and deterministic.

\ifx  not sure how useful this is here...
For example,we can choose $g_{ij} = q(\mathbf{x}_i -\mathbf{x}_j)$ where $q(\cdot)$ is
a multivariate normal density function. Then the candidate $\mathbf{x}_i$ is drawn
by $\mathbf{x}_j + \mathbf{\epsilon}$ where $\mathbf{\epsilon}$ is a random variable.
This is known as random walk \citep{Metropolis1953}.
\fi
In this section, we examine various kinds of connectivity functions and their consequences on the probability flow
dynamics.

\subsection{1-bit flip connections} \label{sec:singleflip}
It can be shown that score matching is a special case of MPF in continuous state spaces, where the connectivity
function is set to connect all states within a small Euclidean distance $r$ in
the limit of $r \rightarrow 0$  \citep{Sohl-Dickstein2011a}.
For simplicity, in the case of a discrete state space (Bernoulli RBM), 
we can fix the Hamming distance to one instead, and consider that data states are connected
to all other states 1-bit flip away:
\begin{equation}
    g_{ij} =
    \begin{cases}
    1, & \text{if state $i,j$ differs by single bit flip} \\
    0, & \text{otherwise}
    \end{cases}
\end{equation}
1-bit flip connectivity gives us a sparse $\Gamma$ with $2^DD$ non-zero terms (rather than a full $2^{2D}$), and may be seen as NPS
where the only negative particles are those which are 1-bit flip away from data states. Therefore, we only ever evaluate $\mathcal{|D|}D$ 
terms from this matrix, making the formulation tractable. This was the only connectivity function
pursued in \citep{Sohl-Dickstein2011a} and is a natural starting point for the approach.

\begin{algorithm}[hpt]
    \caption{Minimum probability flow learning with single bit-flip
      connectivity. Note we leave out all $g_{ij}$ since here we are explicit about
only connecting states of Hamming distance $1$.}\label{mpf_algo}
\begin{itemize}
    \item Initialize the parameters $\theta$ 
	\item {\bf for} each training example $d \in \mathcal{D}$ {\bf do}
\begin{enumerate}
    \item Compute the list of states, $L$, with Hamming distance $1$ from $d$
    \item Compute the probability flow $\Gamma_{id} = \exp{(\frac{1}{2}(F_d(\theta)-F_i(\theta))} $ for each $i \in L$
    \item The cost function for $d$ is $\sum_{i \in L} \Gamma_{id} $ 
    \item Compute the gradient of the cost function, $\frac{\partial J(\theta)}{\partial \theta}  = 
            \sum_{i \in L} \Big( \frac{\partial F_d(\theta)}{\partial \theta} 
                - \frac{\partial F_i(\theta)}{\partial \theta} \Big) \Gamma_{id} $
    \item Update parameters via gradient descent with $\theta \leftarrow \theta  - \lambda \nabla J(\theta)$
\end{enumerate}
{\bf end for}
\end{itemize}
\end{algorithm}
\vspace{-0.5cm}

\subsection{Factorized Minimum Probability Flow}

Previously, we considered connectivity $g_{ij}$ as a binary indicator function of both states $i$ and $j$. Instead, we may wish to use a probability 
distribution, such that $g_{ij}$ is the probability that state $j$ is connected to state $i$ (i.e.~$\sum_i g_{ij} = 1$). 
Following \citep{Sohl-Dickstein2011b}, we simplify this approach by letting $g_{ij} = g_i$, yielding an {\em independence chain} \citep{Tierney1994}. 
This means the probability of being connected to state $i$ is independent of $j$, giving us an alternative way of constructing a transition matrix 
such that the objective function can be factorized:
\begin{align}
    J(\theta) =& \frac{1}{|\mathcal{D}|} \sum_{j\in\mathcal{D}}\sum_{i\not \in \mathcal{D}}
    g_i\left(\frac{g_j}{g_i}\right)^{\frac{1}{2}} \exp \left( \frac{1}{2}\big(F_j(\mathbf{x};\theta) - F_i(\mathbf{x};\theta)\big)\right)
	\label{fmpf_obj0} \\
    =& \left (\frac{1}{|\mathcal{D}|} \sum_{j\in\mathcal{D}} \exp \left( \frac{1}{2} \big( F_j(\mathbf{x};\theta) +\log g_j \big) \right ) \right)
        \left ( \sum_{i\not \in\mathcal{D}} g_i \exp \left( \frac{1}{2} \big(-F_i(\mathbf{x};\theta) +\log g_i\big) \right) \right)
    \label{fmpf_obj}
\end{align}
where $\left(\frac{g_j}{g_i}\right)^{\frac{1}{2}}$ is a scaling term required to counterbalance the difference between $g_i$ and $g_j$.
The independence in the connectivity function allows us to factor all the $j$ terms in \ref{fmpf_obj0} out of the inner sum, leaving us with a product of sums,
something we could not achieve with 1-bit flip connectivity since the connection to state $i$ depends on it being a neighbor of state $j$.
Note that, intuitively, learning is facilitated by connecting data states to states that are probable under the model (i.e.~to contrast the divergence).
Therefore, we can use $p(v;\theta)$ to approximate $g_i$. In practice, for each iteration $n$ of learning, we need the $g_i$ and $g_j$ terms to act as constants with respect to updating $\theta$, and thus we sample them from $p(v;\theta^{n-1})$. We can then rewrite the objective function as
$J(\theta) = J_{\mathcal{D}}(\theta) J_{\mathcal{S}}(\theta)$
\begin{align}
J_{\mathcal{D}}(\theta)=& \left (\frac{1}{|\mathcal{D}|} \sum_{\mathbf{x}\in\mathcal{D}} \exp \left[\frac{1}{2} \big( F(\mathbf{x}; \theta) - F(\mathbf{x};\theta^{n-1})\big)\right]\right);
J_{\mathcal{S}}(\theta)=\left ( \frac{1}{|\mathcal{S}|}\sum_{\mathbf{x}' \in\mathcal{S}} \exp \left[\frac{1}{2} \big(-F(\mathbf{x}'; \theta) + F(\mathbf{x}';\theta^{n-1}) \big) \right]\right) \nonumber
\end{align}
where $\mathcal{S}$ is the sampled set from $p(\mathbf{v};\theta^{n-1})$, and the normalization terms in $\log g_j$ and $\log g_i$ cancel out. 
Note we use the $\theta^{n-1}$ notation to refer to the parameters at the previous iteration, and simply $\theta$ for the current iteration.

\subsection{Persistent Minimum Probability Flow}\label{flow:pmpf}
There are several ways of sampling ``fantasy particles'' from $p(\mathbf{v};\theta^{n-1})$. Notice
that taking the data distribution with respect to $\theta^{n-1}$ is necessary for stable learning. 

Previously, persistent contrastive divergence (PCD) was developed to improve CD-k learning
\citep{Tieleman2009}. Similarly, persistence can be applied to sampling in MPF connectivity functions.
For each update, we pick a new sample based on a MCMC sampler which starts from previous samples.
Then we update $\theta^{n}$, which satsifies $J(\theta^{n}) \leq J(\theta^{n-1})$ \citep{Sohl-Dickstein2011b}.
The pseudo-code for persistent MPF is the same as Factored MPF except for drawing new samples,
which is indicated by square brackets in Algorithm \ref{pmpf_algo}.

As we will show, using persistence in MPF is important for achieving faster convergence in learning.
While the theoretical formulation of MPF guarantees eventual convergence, the focus on minimizing the initial
probability flow will have little effect if the sampler mixes too slowly.
In practice, combining the persistent samples and non-persistent samples gave better performance. 

\begin{algorithm}[hpt]
    \caption{Factored [Persistent] MPF learning with probabilistic connectivity.}\label{pmpf_algo}
\begin{itemize}
    \item {\bf for} each epoch $n$ {\bf do}
    \begin{enumerate}
    	\item Draw a new sample $S^{n}$ based on $S^0$ $\left[S^{n-1}\right]$ using an MCMC sampler.
	\item Compute $J_{\mathcal{S}}(\theta)$
        \item {\bf for} each training example $d \in \mathcal{D}$ {\bf do}
        \begin{enumerate}
            \item Compute $J_d(\theta)$. The cost function for $d$ is $J(\theta)=J_d(\theta)J_{\mathcal{S}}(\theta)$
            \item Compute the gradient of the cost function, \\ $\frac{\partial J(\theta)}{\partial \theta}  = 
                    J_{\mathcal{S}}(\theta) J_d(\theta) \frac{\partial F_d(\theta)}{\partial \theta} + \frac{1}{|\mathcal{S}|} J_{d} \sum_{x' \in \mathcal{S}} \Big(\frac{\partial F(x')}{\partial \theta} \exp \left[\frac{1}{2} \big( F(\mathbf{x'}; \theta) - F(\mathbf{x'};\theta^{n-1})\big)\right] \Big) $
    	    \item Update parameters via gradient descent with $\theta \leftarrow \theta - \lambda \nabla J(\theta)$
        \end{enumerate}
\end{enumerate}
{\bf end for}
\end{itemize}
\end{algorithm}
\vspace{-0.5cm}

\section{Experiments}
We conducted the first empirical study of MPF under different types of
connectivity as discussed in Section \ref{transition_rates}. We  compared our results to
CD-k with varying values for $K$.
We analyzed the MPF variants based on training
RBMs and assessed them quantitatively and qualitatively by comparing the log-liklihoods of the test data and samples
generated from model.
For the experiments, we denote the 1-bit flip, factorized, and persistent methods as MPF-1flip, FMPF, and PMPF, respectively.

The goals of these experiments are to
\vspace{-0.2cm}
\begin{enumerate}
    \setlength{\itemsep}{1pt}
    \setlength{\parskip}{0pt}
    \setlength{\parsep}{0pt}
    \item Compare the performance between MPF algorithms under
      different connectivities; and
    \item Compare the performance between MPF and CD-k.
\end{enumerate}
\vspace{-0.2cm}

In our experiments, we considered the MNIST and CalTech Silhouette datasets.
MNIST consists of 60,000 training and 10,000 test images of size 28 $\times$ 28 pixels containing
handwritten digits from the classes 0 to 9. The pixels in MNIST are binarized based on thresholding.
From the 60,000 training examples, we set aside 10,000 as validation examples to tune the hyperparameters
in our models. The CalTech Silhouette dataset contains the outlines of objects from the CalTech101 dataset, which
are centered and scaled on a 28 $\times$ 28 image plane and rendered as filled black regions on a white
background creating a silhouette of each object. The training set consists of 4,100 examples, with
at least 20 and at most 100 examples in each category.
The remaining instances were split evenly between validation and testing%
\footnote{More details on pre-processing the CalTech Silhouettes can be found in http://people.cs.umass.edu/~marlin/data.shtml}.
Hyperparameters such as learning rate, number of epochs, and batch size were selected
from discrete ranges and chosen based on a held-out validation set.
The learning rate for FMPF and PMPF were chosen from the range [0.001, 0.00001] and 
the learning rate for 1-bit flip was chosen from the range [0.2, 0.001].

\subsection{MNIST - exact log likelihood}
\label{mnist:exact}

In our first experiment, we trained eleven RBMs on the MNIST digits. All RBMs consisted of
20 hidden units and 784 (28$\times$28) visible units.
Due to the small number of hidden variables, we calculated the exact value of the partition function by explicitly
summing over all visible configurations. Five RBMs were learned by PCD1, CD1,
CD10, CD15, and CD25.
Seven RBMs were learned by 1 bit flip, FMPF, and FPMPF\footnote{FPMPF
  is the composition of the FMPF and PMPF connectivities.}.
Block Gibbs sampling is required for FMPF-k and FPMPF-k similar to CD-k training, where the number of steps is given by $k$.

The average log test likelihood values of RBMs with 20 hidden units are presented in Table \ref{sRBM_exp}.
This table gives a sense of the performance under different types of MPF dynamics when the partition function can be calculated exactly.
We observed that PMPF consistently achieved a higher log-likelihood than FMPF.
MPF with 1 bit flip was very fast but gave poor performance compared to FMPF and PMPF.
We also observed that MPF-1flip outperformed CD1.
FMPF always performed slightly worse than CD-k training with the same number of Gibbs steps.
However, PMPF always outperformed CD-k.

\begin{figure*}[t]
    \vspace{-0.3cm}
    \centering
    \includegraphics[width=0.98\textwidth]{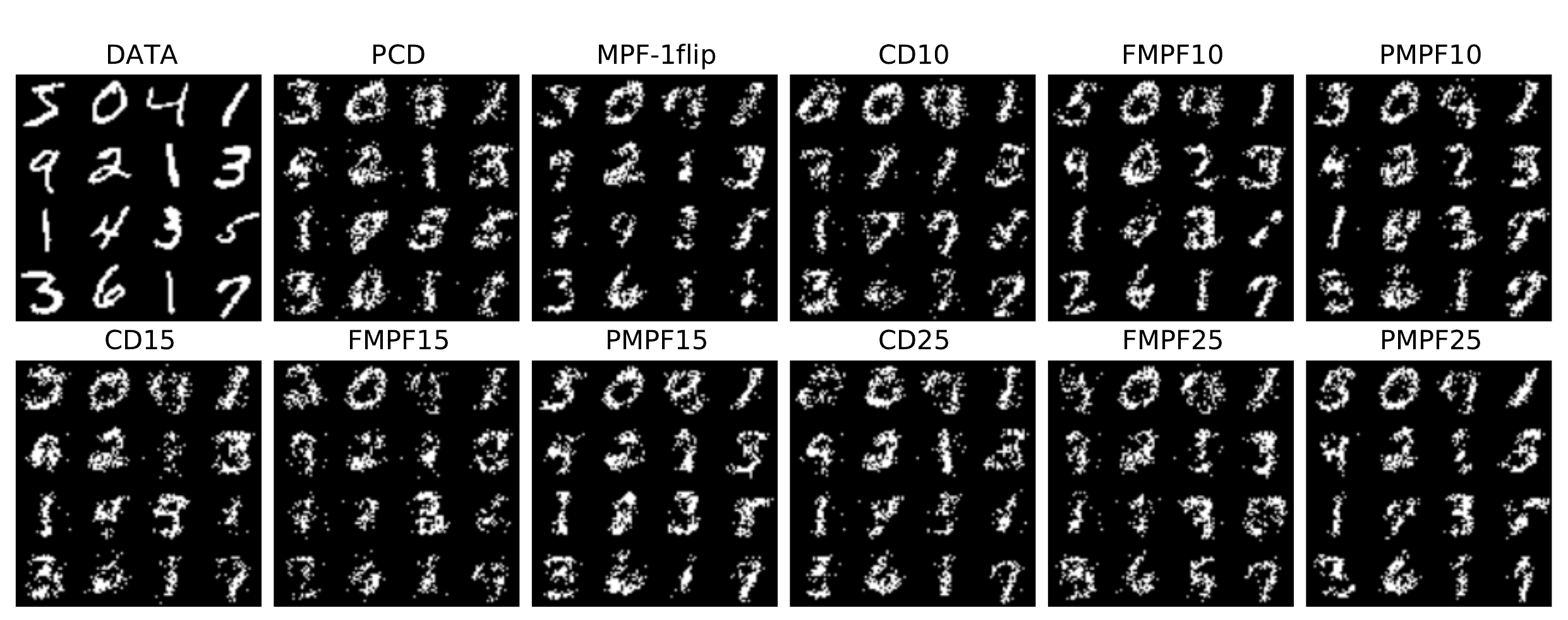}
\caption{Samples generated from the training set. Samples in each panel are
generated by RBMs trained under different paradigms as noted above each image.}
\label{sRBM_figs}
\end{figure*}

One advantage of FMPF is that it converges much quicker than CD-k or PMPF.
This is because we used twice many samples as PMPF
as mentioned in Section \ref{flow:pmpf}.
Figure \ref{sRBM_figs} shows initial data and the generated samples after running 100 Gibbs steps
from each RBM. PMPF produces samples that are visually more appealing than the other methods.

\begin{table*}[t]
\centering
\begin{small}
\caption{Experimental results on MNIST using 11 RBMs with 20 hidden units each. The average
training and test log-probabilities over 10 repeated runs with random parameter initializations are reported.}
\label{sRBM_exp}
\begin{tabular}{lcccr}
\hline
Method & Average log Test & Average log Train & Time (sec)  & Batchsize\\
\hline
CD1    & -145.63 $\pm$ 1.30 & -146.62 $\pm$ 1.72 & 831    & 100\\
PCD    & \bf{-136.10 $\pm$ 1.21} & \bf{-137.13 $\pm$ 1.21} & 2620   & 300\\
MPF-1flip  & -141.13 $\pm$ 2.01 & -143.02 $\pm$ 3.96 & 2931 & 75 \\
\hline
CD10   & -135.40 $\pm$ 1.21 & -136.46 $\pm$ 1.18 & 17329  & 100\\
FMPF10 & -136.37 $\pm$ 0.17 & -137.35 $\pm$ 0.19 & 12533   & 60\\
PMPF10 & -141.36 $\pm$ 0.35 & -142.73 $\pm$ 0.35 & 11445   & 25\\ 
FPMPF10 & \bf{-134.04 $\pm$ 0.12} & \bf{-135.25 $\pm$ 0.11} & 22201  & 25\\
\hline
CD15   & -134.13 $\pm$ 0.82 & -135.20 $\pm$ 0.84 & 26723  & 100\\
FMPF15 & -135.89 $\pm$ 0.19 & -136.93 $\pm$ 0.18 & 18951   & 60\\
PMPF15 & -138.53 $\pm$ 0.23 & -139.71 $\pm$ 0.23 & 13441   & 25\\ 
FPMPF15 & \bf{-133.90 $\pm$ 0.14} & \bf{-135.13 $\pm$ 0.14} & 27302  & 25\\
\hline
CD25   & -133.02 $\pm$ 0.08 & -134.15 $\pm$ 0.08 & 46711  & 100\\
FMPF25 & -134.50 $\pm$ 0.08 & -135.63 $\pm$ 0.07 & 25588  & 60\\
PMPF25 & -135.95 $\pm$ 0.13 & -137.29 $\pm$ 0.13 & 23115  & 25\\ 
FPMPF25 & \bf{-132.74 $\pm$ 0.13} & \bf{-133.50 $\pm$ 0.11} & 50117  & 25\end{tabular}
\end{small}
\vspace{-0.6cm}
\end{table*}

\subsection{MNIST - estimating log likelihood}
In our second set of experiments, we trained RBMs with 200 hidden units. We trained them exactly as described in Section~\ref{mnist:exact}.
These RBMs are able to generate much higher-quality samples from the data distribution, however, the partition function
can no longer be computed exactly.

In order to evaluate the model quantitatively, we estimated the test log-likelihood
using the Conservative Sampling-based Likelihood estimator (CSL) \citep{Bengio2013} and
annealed importance sampling (AIS) \citep{Salakhutdinov2008}.
Given well-defined conditional probabilities $P(\mathbf{v}|\mathbf{h})$ of a model and
a set of latent variable samples $S$ collected from a Markov chain, CSL computes
\vspace{-0.1cm}
\begin{equation}
    \log \hat{f}(\mathbf{v}) = \log \text{mean}_{h\in S} P(\mathbf{v}|\mathbf{h}).
    \vspace{-0.1cm}
\end{equation}
The advantage of CSL is that sampling latent variables $\mathbf{h}$ instead of $\mathbf{v}$
has the effect of reducing the variance of the estimator. Also, in contrast to annealed importance sampling (AIS)
\citep{Salakhutdinov2008}, which tends to overestimate, CSL is much more conservative in its estimates.
However, most of the time, CSL is far off from the true estimator, so we bound our negative log-likelihood
estimate from above and below using both AIS and CSL. 

\begin{figure*}[t]
    \vspace{-0.3cm}
    \centering
    \includegraphics[width=0.98\textwidth]{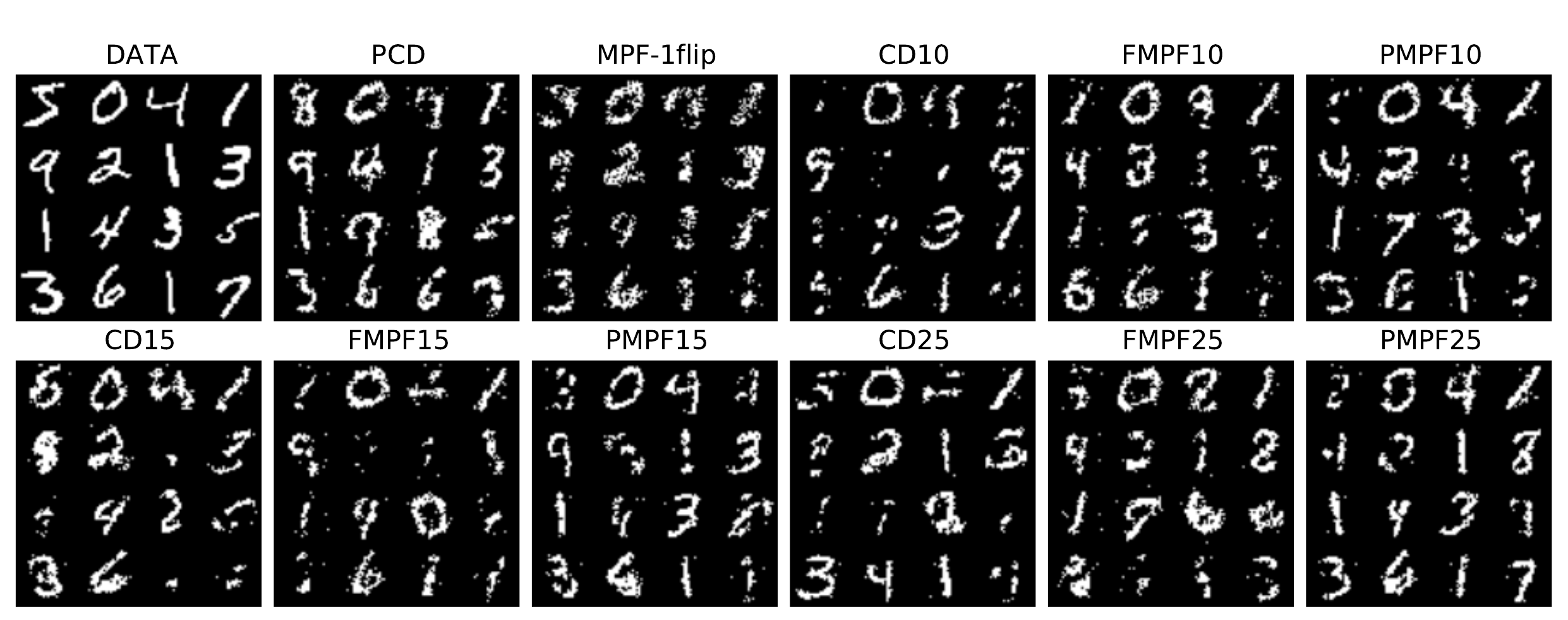}
\caption{Samples generated from the training set. Samples in each panel are
generated by RBMs trained under different paradigms as noted above each image.}
\label{fig:RBM_mnist}
\end{figure*}
\begin{table*}[t]
\centering
\begin{small}
\caption{Experimental results on MNIST using 11 RBMs with 200 hidden units each. The average estimated
training and test log-probabilities over 10 repeated runs with random parameter initializations are reported. Likelihood estimates are made with CSL \citep{Bengio2013}
and AIS \citep{Salakhutdinov2008}}
\label{tab:rbm_mnist}
\begin{tabular}{lcccccr}
& \multicolumn{2}{c}{CSL} &\multicolumn{2}{c}{AIS} &  \multicolumn{2}{c}{} \\ \hline
Method & Avg. log Test & Avg. log Train & Avg. log Test & Avg. log Train & Time (sec)  & Batchsize \\
\hline\hline
 \noalign{\global\arrayrulewidth0.4pt}
CD1    & -138.63 $\pm$ 0.48 & -138.70 $\pm$ 0.45 & -98.75 $\pm$ 0.66 & -98.61 $\pm$ 0.66 & 1258   & 100\\ 
PCD1   & \bf{-114.14 $\pm$ 0.26} & \bf{-114.13 $\pm$ 0.28} & \bf{-88.82 $\pm$ 0.53} & \bf{-89.92 $\pm$ 0.54} & 2614   & 100\\ 
MPF-1flip    & -179.73 $\pm$ 0.085 & -179.60 $\pm$ 0.07 & -141.95 $\pm$ 0.23 & -142.38 $\pm$ 0.74 & 4575 & 75\\
\hline
CD10   & -117.74 $\pm$ 0.14 & -117.76 $\pm$ 0.13    & -91.94 $\pm$ 0.42 & -92.46 $\pm$ 0.38 & 24948  & 100\\
FMPF10 & -115.11 $\pm$ 0.09 & -115.10 $\pm$ 0.07    & -91.21 $\pm$ 0.17 & -91.39 $\pm$ 0.16 & 24849  & 25\\
PMPF10 & -114.00 $\pm$ 0.08 & -113.98 $\pm$ 0.09    & -89.26 $\pm$ 0.13 & -89.37 $\pm$ 0.13 & 24179  & 25\\
FPMPF10& \bf{-112.45 $\pm$ 0.03} & \bf{-112.45 $\pm$ 0.03}    & \bf{-83.83 $\pm$ 0.23} & \bf{-83.26 $\pm$ 0.23} & 24354  & 25\\
\hline
CD15   & -115.96 $\pm$ 0.12 & -115.21 $\pm$ 0.12    & -91.32 $\pm$ 0.24 & -91.87 $\pm$ 0.21 & 39003  & 100\\
FMPF15 & -114.05 $\pm$ 0.05 & -114.06 $\pm$ 0.05    & -90.72 $\pm$ 0.18 & -90.93 $\pm$ 0.20 & 26059  & 25\\
PMPF15 & -114.02 $\pm$ 0.11 & -114.03 $\pm$ 0.09    & -89.25 $\pm$ 0.17 & -89.85 $\pm$ 0.19 & 26272 & 25\\
FPMPF15& \bf{-112.58 $\pm$ 0.03} & \bf{-112.60 $\pm$ 0.02}    & \bf{-83.27 $\pm$ 0.15} & \bf{-83.84 $\pm$ 0.13} & 26900  & 25\\
\hline
CD25   & -114.50 $\pm$ 0.10 & -114.51 $\pm$ 0.10    & -91.36 $\pm$ 0.26 & -91.04 $\pm$ 0.25 & 55688  & 100\\
FMPF25 & -113.07 $\pm$ 0.06 & -113.07 $\pm$ 0.07    & -90.43 $\pm$ 0.28 & -90.63 $\pm$ 0.27 & 40047  & 25 \\
PMPF25  & -113.70 $\pm$ 0.04 & -113.69 $\pm$ 0.04   & -89.21 $\pm$ 0.14 & -89.79 $\pm$ 0.13 & 52638  & 25 \\
FPMPF25 & \bf{-112.38 $\pm$ 0.02} & \bf{-112.42 $\pm$ 0.02}   & \bf{-83.25 $\pm$ 0.27} & \bf{-83.81 $\pm$ 0.28} & 53379  & 25 \\

\end{tabular}
\end{small}
\vspace{-0.6cm}
\end{table*}

Table \ref{tab:rbm_mnist} demonstrates the test log-likelihood of various RBMs with 200 hidden units.
The ranking of the different training paradigms with respect to performance was similar to what we observed in Section~\ref{mnist:exact} with PMPF emerging as the winner.
However, contrary to the first experiment, we observed that MPF with 1 bit flip did
not perform well. 
Moreover, FMPF and PMPF both tended to give higher test log-likelihoods than CD-k training.
Smaller batch sizes worked better with MPF when the number of hiddens was increased.
Once again, we observed smaller variances compared to CD-k with both forms of MPF, especially with FMPF.
We noted that FMPF and PMPF always have smaller variance compared to CD-k. This implies that FMPF and PMPF are
less sensitive to random weight initialization.
Figure \ref{fig:RBM_mnist} shows initial data and generated samples after running 100 Gibbs steps
for each RBM. PMPF clearly produces samples that look more like digits.

\begin{figure*}[t]
    \vspace{-0.3cm}
    \centering
    \includegraphics[width=1.0\textwidth]{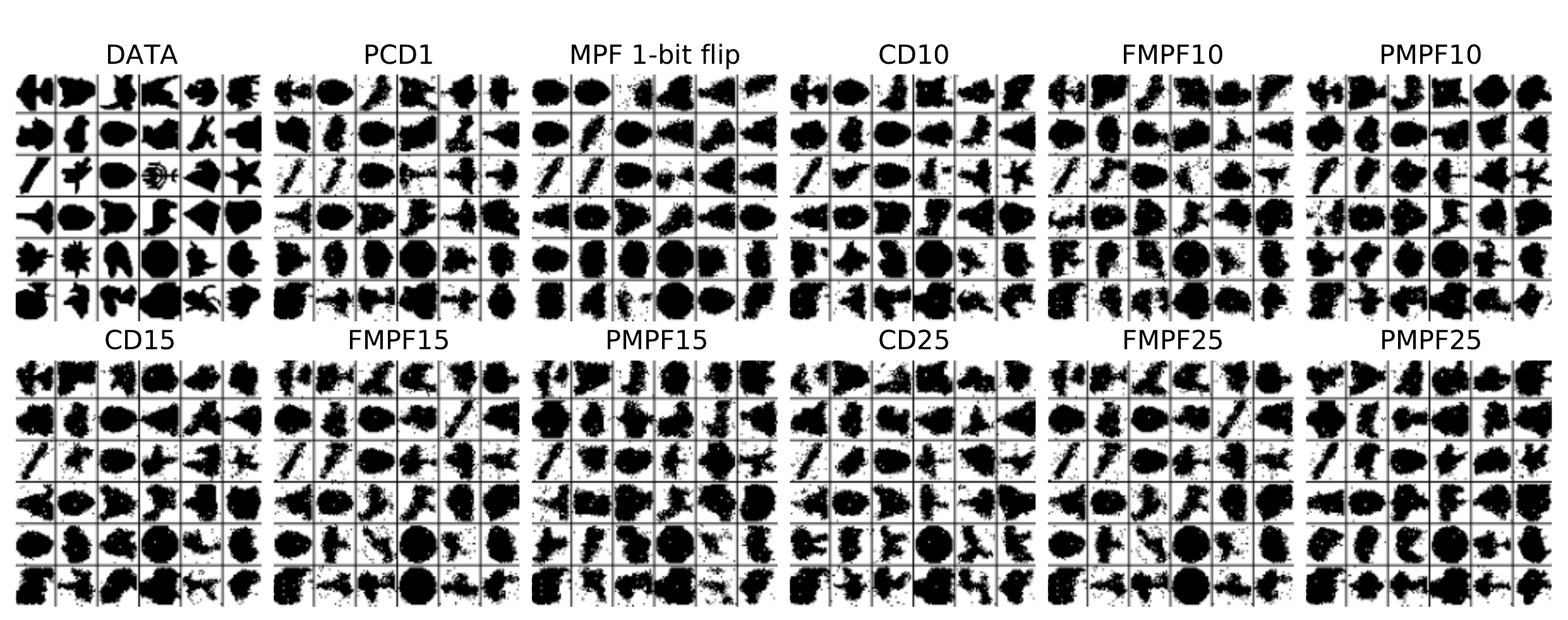}
    \caption{Random samples generated by RBMs with different training procedures.}
    \label{fig:RBM_caltech28}
\end{figure*}
\begin{table*}[t]
\centering
\begin{small}
\caption{Experimental results on Caltech-101 Silhouettes using 11 RBMs with 500 hidden units each. The average estimated
training and test log-probabilities over 10 repeated runs with random parameter initializations are reported. Likelihood estimates are made with CSL \citep{Bengio2013}
and AIS \citep{Salakhutdinov2008}.}
\label{tab:caltech}
\begin{tabular}{lcccccr}
\hline
& \multicolumn{2}{c}{CSL} &\multicolumn{2}{c}{AIS} &  \multicolumn{2}{c}{} \\ \hline
Method & Avg. log Test & Avg. log Train & Avg. log Test & Avg. log Train & Time (sec)  & Batchsize\\
\hline\hline
CD1          & -251.30 $\pm$ 1.80 & -252.04 $\pm$ 1.56   & -141.87 $\pm$ 8.80 & -142.88 $\pm$ 8.85 & 300   & 100\\
PCD1         & \bf{-199.89 $\pm$ 1.53} & \bf{-199.95 $\pm$ 1.31}   & \bf{-124.56 $\pm$ 0.24} & \bf{-116.56 $\pm$ 2.40} & 784   & 100\\
MPF-1flip    & -281.55 $\pm$ 1.68 & -283.03 $\pm$ 0.60   & -164.96 $\pm$ 0.23 & -170.92 $\pm$ 0.20 & 505   & 100\\
\hline
CD10         & -207.77 $\pm$ 0.92 & -207.16 $\pm$ 1.18   & -128.17 $\pm$ 0.20 & -120.65 $\pm$ 0.19 & 4223  & 100\\
FMPF10       & -211.30 $\pm$ 0.84 & -211.39 $\pm$ 0.90   & -135.59 $\pm$ 0.16 & -135.57 $\pm$ 0.18 & 2698  & 20\\
PMPF10      & -203.13 $\pm$ 0.12 & -203.14 $\pm$ 0.10   & -128.85 $\pm$ 0.15 & -123.06 $\pm$ 0.15  &  7610 & 20\\
FPMPF10      & \bf{-200.36 $\pm$ 0.16} & \bf{-200.16 $\pm$ 0.16}   & \bf{-123.35 $\pm$ 0.16} & \bf{-108.81 $\pm$ 0.15} & 11973  & 20\\
\hline
CD15         & -205.12 $\pm$ 0.87 & -204.87 $\pm$ 1.13   & -125.08 $\pm$ 0.24 & -117.09 $\pm$ 0.21 & 6611  & 100\\
FMPF15       & -210.66 $\pm$ 0.24 & -210.19 $\pm$ 0.30   & -130.28 $\pm$ 0.14 & -128.57 $\pm$ 0.15 & 3297  & 20\\
PMPF15       & -201.47 $\pm$ 0.13 & -201.67 $\pm$ 0.10 & -127.09 $\pm$ 0.10 & -121 $\pm$ 0.12 & 9603  & 20\\
FPMPF15      & \bf{-198.59 $\pm$ 0.17} & \bf{-198.66 $\pm$ 0.17}   & \bf{-122.33 $\pm$ 0.13} & \bf{-107.88 $\pm$ 0.14} & 18170  & 20\\
\hline
CD25         & -201.56 $\pm$ 0.11 & -201.50 $\pm$ 0.13   & -124.80 $\pm$ 0.20 & -117.51 $\pm$ 0.23 & 13745 & 100\\
FMPF25       & -206.93 $\pm$ 0.13 & -206.86 $\pm$ 0.11  & -129.96 $\pm$ 0.07 & -127.15 $\pm$ 0.07 & 10542 & 10\\
PMPF25       &  -199.53 $\pm$ 0.11 & -199.51 $\pm$ 0.12   & -127.81 $\pm$ 020 & -122.23 $\pm$ 0.17 & 18550 & 10\\
FPMPF25      & \bf{-198.39 $\pm$ 0.0.16} & \bf{-198.39 $\pm$ 0.17}   & \bf{-122.75 $\pm$ 0.13} & \bf{-108.32 $\pm$ 0.12} & 23998 & 10\\

\end{tabular}
\end{small}
\vspace{-0.6cm}
\end{table*}

\subsection{Caltech 101 Silhouettes - estimating log likelihood}
Finally, we evaluated the same set of RBMs on the Caltech-101 Silhouettes dataset. Compared to MNIST, this dataset
contains much more diverse structures with richer correlation among the pixels.
It has 10 times more categories, contains less training data per category,
and each object covers more of the image. For these reasons, we use
500 hidden units per RBM. The estimated average log-likelihood of train
and test data is presented in Table \ref{tab:caltech}.\looseness=-1

The results for Caltech 101 Silhouettes are consistent with MNIST. In
every case, we observed a larger margin between PMPF and CD-k when the
number of sampling steps was smaller.
In addition, the single bit flip technique was not particularly successful, especially
as the number of latent variables grew. We speculate that the reason for this might have
 to do with the slow rate of convergence for the dynamic system. Moreover, PMPF works better than FMPF for
 similar reasons. By having persistent samples as the learning progresses,
the dynamics always begin closer to equilibrium, and hence converge more quickly.
Figure \ref{fig:RBM_caltech28} shows initial data and generated samples after running 100 Gibbs steps
for each RBM on Caltech28 dataset.

\section{Conclusion}
MPF is an unsupervised learning algorithm that can be
employed off-the-shelf to any energy-based model.
It has a number of favorable properties but has not seen application
proportional to its potential.
In this paper, we first expounded on MPF and its connections to CD-k training,
which allowed us to gain a better understanding and perspective to CD-k.
We proved a general form for the transition matrix such that the
equilibrium distribution converges to that of an RBM. This may lead
to future extensions of MPF based on the choice of $o(\cdot)$ in Equation \ref{eqn:new_gamma}.\looseness=-1

One of the merits of MPF is that the choice of designing a dynamic system by
defining a connectivity function is left open as long as it satisfies the fixed point equation.
We thoroughly explored three different connectivity structures, noting that connectivity
can be designed inductively or by sampling. Finally, we showed empirically that
MPF, and in particular, PMPF, outperforms CD-k for training generative models.
Until now, RBM training was dominated by methods based on CD-k; however, our results
indicate that MPF is a practical and effective alternative.

\clearpage

\bibliography{mpf_rbm}

\begin{thebibliography}{14}
\providecommand{\natexlab}[1]{#1}
\providecommand{\url}[1]{\texttt{#1}}
\expandafter\ifx\csname urlstyle\endcsname\relax
  \providecommand{\doi}[1]{doi: #1}\else
  \providecommand{\doi}{doi: \begingroup \urlstyle{rm}\Url}\fi

\bibitem[Bastien et~al.(2012)Bastien, Lamblin, Pascanu, Bergstra, Goodfellow,
  Bergeron, Bouchard, and Bengio]{Theano}
Bastien, Fr{\'{e}}d{\'{e}}ric, Lamblin, Pascal, Pascanu, Razvan, Bergstra,
  James, Goodfellow, Ian~J., Bergeron, Arnaud, Bouchard, Nicolas, and Bengio,
  Yoshua.
\newblock Theano: new features and speed improvements.
\newblock Deep Learning and Unsupervised Feature Learning NIPS 2012 Workshop,
  2012.

\bibitem[Bengio et~al.(2013)Bengio, Yao, and Cho]{Bengio2013}
Bengio, Yoshua, Yao, Li, and Cho, Kyunghyun.
\newblock Bounding the test log-likelihood of generative models.
\newblock In \emph{Proceedings of the International Conference of Learning
  Representations (ICLR)}, 2013.

\bibitem[Besag(1975)]{Besag1975}
Besag, Julian.
\newblock Statistical analysis of non-lattice data.
\newblock \emph{The Statistician}, 24:\penalty0 179--195, 1975.

\bibitem[Hinton(2002)]{Hinton2002}
Hinton, Geoffrey.~E.
\newblock Training products of experts by minimizing contrastive divergence.
\newblock \emph{Neural Computation}, 14:\penalty0 1771--1880, 2002.

\bibitem[Hyv{\"a}rinen(2005)]{Aapo2005}
Hyv{\"a}rinen, Aapo.
\newblock Estimation of non-normalized statistical models by score matching.
\newblock \emph{Journal of Machine Learning Research}, 6:\penalty0 695--709,
  2005.

\bibitem[MacKay(2001)]{MacKay2001}
MacKay, David J.~C.
\newblock Failures of the one-step learning algorithm, 2001.
\newblock URL
  \url{http://www.inference.phy.cam.ac.uk/mackay/abstracts/gbm.html}.
\newblock Unpublished Technical Report.

\bibitem[Marlin \& Freitas(2011)Marlin and Freitas]{Marlin2011}
Marlin, Benjamin~M. and Freitas, Nando~de.
\newblock Asymptotic efficiency of deterministic estimators for discrete
  energy-based models: Ratio matching and pseudolikelihood.
\newblock In \emph{Proceedings of the Uncertainty in Artificial Intelligence
  (UAI)}, 2011.

\bibitem[Salakhutdinov \& Murray(2008)Salakhutdinov and
  Murray]{Salakhutdinov2008}
Salakhutdinov, Ruslan and Murray, Iain.
\newblock On the quantitative analysis of deep belief networks.
\newblock In \emph{Proceedings of the International Conference of Machine
  Learning (ICML)}, 2008.

\bibitem[Smolensky(1986)]{smolensky1986}
Smolensky, Paul.
\newblock Information processing in dynamical systems: Foundations of harmony
  theory.
\newblock In \emph{Parallel Distributed Processing: Volume 1: Foundations},
  pp.\  194--281. MIT Press, 1986.

\bibitem[Sohl-Dickstein(2011)]{Sohl-Dickstein2011b}
Sohl-Dickstein, Jascha.
\newblock Persistent minimum probability flow.
\newblock Technical report, Redwood Centre for Theoretical Neuroscience, 2011.

\bibitem[Sohl-Dickstein et~al.(2011)Sohl-Dickstein, Battaglino, and
  DeWeese]{Sohl-Dickstein2011a}
Sohl-Dickstein, Jascha, Battaglino, Peter, and DeWeese, Michael~R.
\newblock Minimum probability flow learning.
\newblock In \emph{Proceedings of the International Conference of Machine
  Learning (ICML)}, 2011.

\bibitem[Sutskever \& Tieleman(2009)Sutskever and Tieleman]{Sutskever2010}
Sutskever, Ilya and Tieleman, Tijmen.
\newblock On the convergence properties of contrastive divergence.
\newblock In \emph{Proceedings of the AI \& Statistics (AI STAT)}, 2009.

\bibitem[Tieleman \& Hinton(2009)Tieleman and Hinton]{Tieleman2009}
Tieleman, Tijmen and Hinton, Geoffrey~E.
\newblock Using fast weights to improve persistent contrastive divergence.
\newblock In \emph{Proceedings of the International Conference of Machine
  Learning (ICML)}, 2009.

\bibitem[Tierney(1994)]{Tierney1994}
Tierney, Luke.
\newblock Markov chains for exploring posterior distributions.
\newblock \emph{Annals of Statistics}, 22:\penalty0 1701--1762, 1994.

\end{thebibliography}
\bibliographystyle{iclr2015}

\newpage
\appendix
\section{Minimum Probability Flow}
\subsection{Dynamics of The Model} \label{app:appendexA}
\begin{customthm}{1}
    Suppose $p_j^{(\infty)}$ is the probability of state $j$ and $p_i^{(\infty)}$ is the probability of
    state $i$. Let the transition matrix be 
    \begin{equation}
        \Gamma_{ij} = g_{ij}\exp \left( \frac{o(F_i-F_j) +1}{2} (F_j-F_i) \right)
    \end{equation}
    such that $o(\cdot)$ is any odd function, where $g_{ij}$ is the symmetric connectivity between the states $i$ and $j$.
    Then this transition matrix satisfies detailed balance in Equation \ref{detailed_balance}.
\end{customthm}
\begin{proof}
    By cancalling out the partition function, the detailed balance Equation \ref{detailed_balance} can be formulated to be
\begin{equation}
    \Gamma_{ji} \exp{(-F_i)} = \Gamma_{ij}\exp{(-F_j)}
    \label{detailed_balance}
\end{equation}
where $F_i = F(\mathbf{v}=i;\theta)$ 
    We substitute transition matrix defined in Equation \ref{eqn:new_gamma}, then we get the following after straight forward formula 
    manipulation.
    \begin{align}
        &\Gamma_{ji} \exp{(-F_i)} / \Gamma_{ij}\exp{(-F_j)}) = 1\nonumber\\
        &\exp \left( \frac{o(F_i-F_j) +1}{2} (F_j-F_i) - F_i \right) / \exp \left( \frac{o(F_j-F_i) +1}{2} (F_i-F_j) -F_j \right) = 1\nonumber\\
        &\exp \left( \frac{o(F_i-F_j) +1}{2} (F_j-F_i) - F_i  - \frac{o(F_j-F_i) +1}{2} (F_i-F_j)  + F_j \right) = 1\nonumber\\
        &\frac{o(F_i-F_j) +1}{2} (F_j-F_i) - F_i  - \frac{o(F_j-F_i) +1}{2} (F_i-F_j)  + F_j = 0\nonumber\\
        &(F_i-F_j) \bigg( \frac{o(F_i-F_j) +1}{2} + \frac{o(F_j-F_i) +1}{2} - 1\bigg) = 0\nonumber \\
        &(F_i-F_j) \bigg( \frac{o(F_i-F_j) }{2} + \frac{o(F_j-F_i)}{2} \bigg) = 0 \nonumber
    \end{align}
    Notice that since $o(\cdot)$ is an odd function, this makes the term $\big( \frac{o(F_i-F_j) }{2} + \frac{o(F_j-F_i)}{2}\big)=0$. 
    Therefore, the detailed balance criterion is satisfied.
\end{proof}

\end{document}